\def\cvprPaperID{xxxx}
\def\confName{xxxx}
\def\confYear{xxxx}

\def\paperTitle{Sketch Video Synthesis}

\def\authorBlock{
Yudian Zheng\\
University of Saarland
\and
Xiaodong Cun\footnotemark[2]\\
Tencent AI Lab
\and 
Menghan Xia\\
Tencent AI Lab
\and 
Chi-Man Pun\\
University of Macau
\and
\\
\url{https://sketchvideo.github.io}
}

\newif\ifreview 
\newif\ifarxiv \newcommand{\arxiv}{\arxivtrue}
\newif\ifcamera 
\newif\ifrebuttal

\arxiv 

\pdfoutput=1
\documentclass[10pt,twocolumn,letterpaper]{article}
\ifreview \usepackage[review]{cvpr} \fi
\ifarxiv \usepackage[pagenumbers]{cvpr} \fi
\ifrebuttal \usepackage[rebuttal]{cvpr} \fi
\ifcamera \usepackage{cvpr} \fi

\usepackage{graphicx}
\usepackage{amsmath}
\usepackage{amssymb}
\usepackage{booktabs}

\usepackage{times}
\usepackage{microtype}
\usepackage{epsfig}
\usepackage[table,xcdraw]{xcolor}
\usepackage{caption}
\usepackage{float}
\usepackage{placeins}
\usepackage{color, colortbl}
\usepackage{stfloats}
\usepackage{enumitem}
\usepackage{tabularx}
\usepackage{xstring}
\usepackage{multirow}
\usepackage{xspace}
\usepackage{url}
\usepackage{subcaption}
\usepackage{xcolor}
\usepackage[hang,flushmargin]{footmisc}

\usepackage{makecell}

\usepackage{algorithm}
\usepackage{algorithmicx}
\usepackage{algpseudocode}
\usepackage{balance}
\usepackage{color}
\usepackage{listings}
\usepackage{array}
\usepackage{bbding}
\usepackage{algpseudocode}
\ifcamera \usepackage[accsupp]{axessibility} \fi

% ...

\definecolor{colorx}{rgb}{0.92,0.49,0.19}

\newcommand{\RM}[1]{}

\ifarxiv  \fi

\newcommand{\R}[1]{{%
    \textbf{%
        \ifstrequal{#1}{1}{\textcolor{red}{R#1}}{%
        \ifstrequal{#1}{2}{\textcolor{blue}{R#1}}{%
        \ifstrequal{#1}{3}{\textcolor{magenta}{R#1}}{%
        \ifstrequal{#1}{4}{\textcolor{teal}{R#1}}{%
                           \textcolor{cyan}{R#1}%
        }}}}%
    }%
}}

\usepackage{tikz}

\usepackage{xr-hyper}

\makeatletter
\newcommand*{\addFileDependency}[1]{
  \typeout{(#1)}
  \@addtofilelist{#1}
  \IfFileExists{#1}{}{\typeout{No file #1.}}
}

\makeatother

\usepackage[pagebackref,breaklinks,colorlinks]{hyperref}
\usepackage[capitalize]{cleveref}
\crefname{section}{Sec.}{Secs.}
\crefname{table}{Table}{Tables}
\crefname{figure}{Fig.}{Figs.}

\definecolor{alizarin}{rgb}{0.82, 0.1, 0.26}

\hypersetup{
    colorlinks = true,
    linkbordercolor = {white},
    citecolor=blue,
    urlcolor=alizarin
}

\begin{document}
\title{\paperTitle}
\author{\authorBlock}

\twocolumn[{
\maketitle
\begin{center}
    \captionsetup{type=figure}
    \vspace{-1em}
\newcommand{\imwidth}{1.0\textwidth}

\begin{tabular}{@{}c@{}}
\parbox{\imwidth}{\includegraphics[width=\imwidth, ]{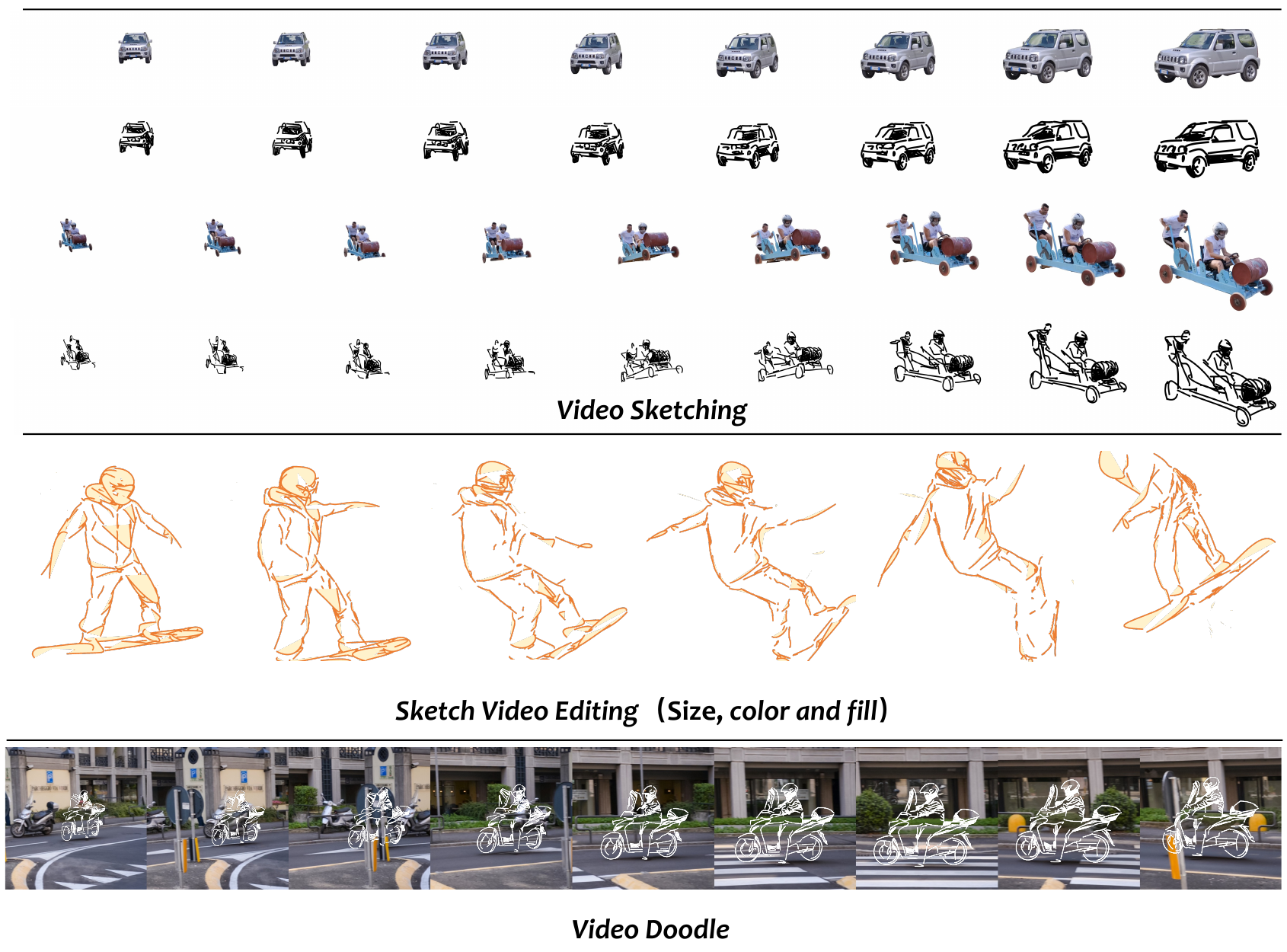}}
\end{tabular}
    \vspace{-1.em}
    \captionof{figure}{Given an input video (of the foreground object), we introduce a novel method for sketching the video using Bézier Curves so that the video can be represented by scalable vector graphics~(SVG). The generated sketch video maintains semantic alignment with the input and exhibits temporal consistency. The flexibility of SVG allows for various rendering techniques, including resizing, color filling, and overlaying doodles on the original background images, enabling the creation of diverse artistic effects.}
    \label{fig:teaser}
\end{center}
}]

\renewcommand{\thefootnote}{\fnsymbol{footnote}}
\footnotetext[2]{~Corresponding author.}


\begin{abstract}
Understanding semantic intricacies and high-level concepts is essential in image sketch generation, and this challenge becomes even more formidable when applied to the domain of videos. To address this, we propose a novel optimization-based framework for sketching videos represented by the frame-wise Bézier Curves. In detail, we first propose a cross-frame stroke initialization approach to warm up the location and the width of each curve. Then, we optimize the locations of these curves by utilizing a semantic loss based on CLIP features~\cite{clip} and a newly designed consistency loss using the self-decomposed 2D atlas network~\cite{layer-atlas}. Built upon these design elements, the resulting sketch video showcases impressive visual abstraction and temporal coherence. Furthermore, by transforming a video into SVG lines through the sketching process, our method unlocks applications in sketch-based video editing and video doodling, enabled through video composition, as exemplified in the teaser.
\end{abstract}
\section{Introduction}

Freehand drawing is a widely adopted method for quickly prototyping ideas across various domains~\cite{opensketch, dl-free-hand-sketch}. This approach embodies simplicity, abstraction, and adaptability, empowering individuals to effectively express their concepts. Furthermore, skilled artists can develop distinctive artistic styles through freehand drawing. While sketching is a common practice for images, often using formats like scalable vector graphics~(SVG), there has been limited exploration of its application in the context of sketching videos. This uncharted territory makes the exploration of sketch videos an intriguing and meaningful endeavor.

While traditional approaches, such as edge detection methods~\cite{hed, canny} excel in rendering realistic sketches, they struggle to create more expressive and abstract representations due to their reliance on mathematical and geometric operations. In an attempt to incorporate semantic awareness, previous sketching methods have attempted to learn human-like sketches from a collected dataset in different levels of abstraction and styles~\cite{10.1145/2461912.2461964, li2019photo, kampelmuhler2020synthesizing} at the pixel level. While these data-driven methods imitate human sketches, the requirements and quality of the relevant datasets restrict the output. As introduced by recent image sketching works~\cite{clipasso, clipascene}, line drawings are defined using the control points of Bézier Curves and are optimized to represent the scene. These methods employ multi-scale deep perceptual losses~\cite{clip} to bridge the gap between generated sketches and real scenes, bypassing the constraints of traditional datasets and yielding diverse results. We follow these frameworks to represent video sketches in SVG format. However, if we simply apply image-based sketching~\cite{clipasso} in a frame-wise manner without careful consideration, the strokes will converge into local minima rapidly. Additionally, the flickering issue of generated video is not easily resolved through conventional video deflickering algorithms~\cite{dvp, all-in-one-defliker}, especially when dealing with vector graphics.

To overcome the challenges mentioned above, we introduce an innovative optimization-based framework aimed at generating sketch videos in SVG format that exhibit both semantic alignment and temporal consistency. To achieve this goal, we leverage Neural Layered Atlas (NLA)~\cite{layer-atlas} to our tasks for multiple purposes. 
NLA is first proposed for video editing, it maps each point in video to a uniform global UV map, so that it can guarantee the correspondences across frames and help for temporal point consistency. 
In detail, the process of generating a high-quality video sketch involves careful initialization and continuous optimization of the sketch video. 
We begin by carefully selecting the initial locations for candidate points, where an effective initialization is crucial for avoiding unfavorable local minima~\cite{clipasso} and accurately conveying the video's semantics.
To achieve this, our initialization approach utilizes semantic-aware edges, derived from salient maps obtained from the combination of CLIP features~\cite{clip} and XDoG edge detection~\cite{xdog}.
Subsequently, these selected points are propagated to all frames and optimized to their initial positions using a pre-trained NLA. 
Then, we optimize the location of these points using several losses so that they can ensure both semantic alignment and temporal consistency. We transform the candidate points to Bézier Curves and utilize a differentiable rasterizer~\cite{diffvg} to render them to the frame-wise images. For semantic abstraction, we utilize the pre-trained CLIP image encoder as a feature extractor and compute losses between the rendered one and the real video frame. For temporal consistency, we ensure consistency of vector points via the pre-trained NLA~\cite{layer-atlas} so that they can be consistent from a global view. Based on these techniques, the proposed method can successfully generate the abstraction sketches of the specific given video.

In addition to streamlining the process of sketching videos, our approach paves the way for various video applications. For instance, it allows for the creation of colorful videos by applying drawing techniques to a single frame. Furthermore, our method introduces novel possibilities in video editing, such as substituting the original content by integrating sketches into the scenes. Additionally, our approach enables the generation of video doodles to enhance other video content. 

The contribution of this paper can be summarized as:

\begin{itemize}
\item We first tackle the problem of generating scalable abstract videos via several Bézier curves.
\item By utilizing the consistency of pre-trained video implicit representation~\cite{layer-atlas}, we propose a novel point initialization method and a temporal consistency loss for video sketching synthesis.
\item Our method generates the animated SVG from the input video, enabling multiple new applications of video editing and doodles.
\end{itemize}
\section{Related Work}

\subsection{Doodling and Abstraction}
Doodling and abstraction are common forms of artistic expression that use few and sparse curves to depict objects or scenes abstractly. Early works in sketching, such as traditional edge detection methods~\cite{canny, hed}, effectively describe the structure and semantics of images. While these methods produce clear and coherent sketch videos, they lean towards excessive realism, often neglecting the artistic preferences of viewers. Combining edge detection methods with line drawing video stylization~\cite{LineDrawing} generates abstract images, but it often overlooks the holistic aspect of video frames and loses video consistency. Recent data-driven methods primarily focus on generating human-like sketches through domain adaptation using carefully curated datasets~\cite{Ha_Eck_2017, Arbeláez_Maire_Fowlkes_Malik_2010}. These methods offer the ability to create various styles and levels of abstraction. However, the resulting style is often closely tied to the specific dataset characteristics, making it less suitable for unstructured and previously unseen data. Extending these methods to handle videos, which are typically supervised by image datasets, is also a challenging endeavor. Another method, such as video doodling~\cite{VideoDoodles}, is used in sketch addition on video, but it is not satisfied with a global level of abstraction.

\begin{figure*}[t]
    \centering
    \includegraphics[width=0.99\textwidth]{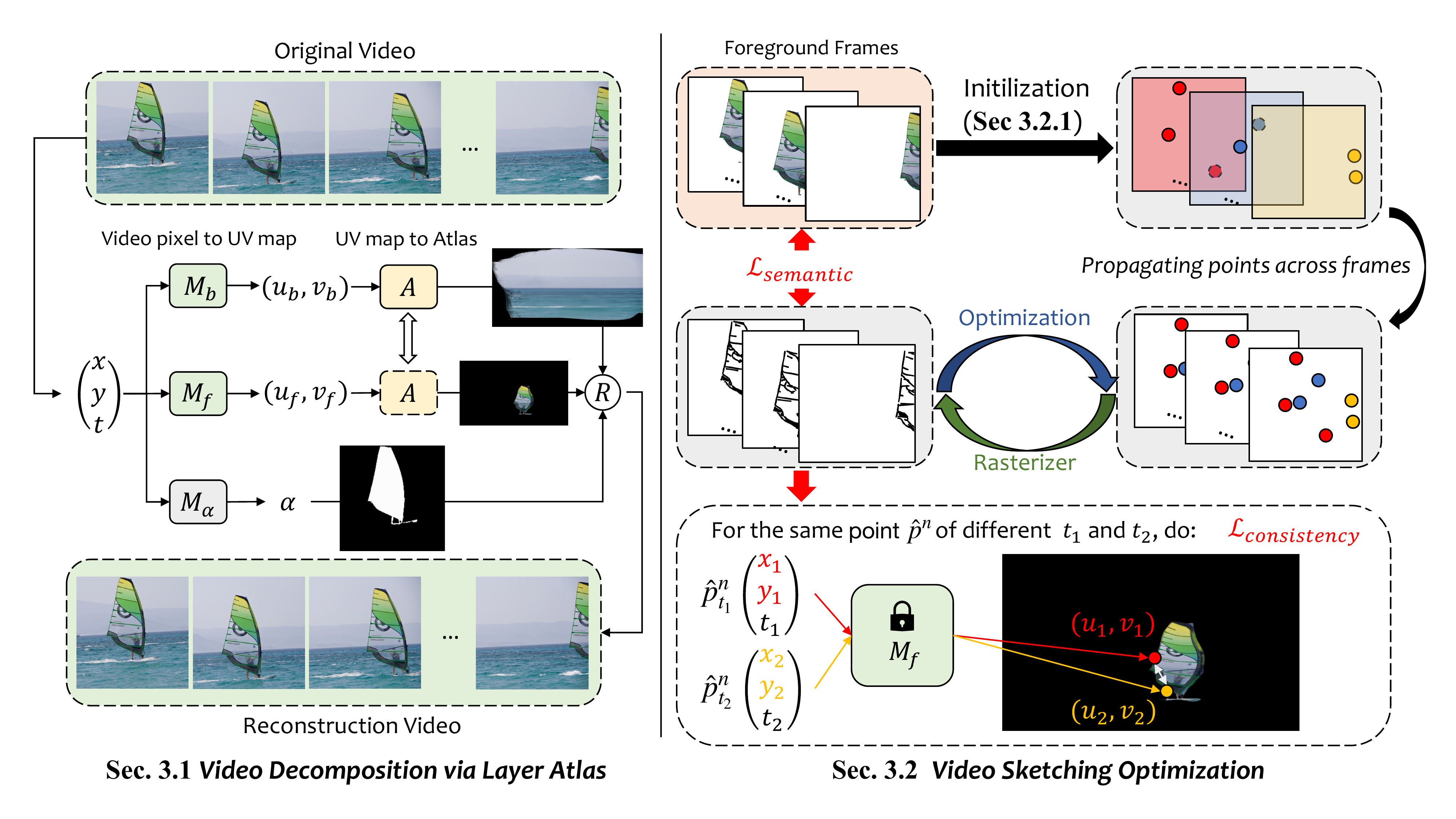}
    \vspace{-1em}
    \caption{\textbf{The pipeline of the whole framework.} Firstly, we train a layer atlas to decompose the video into the trained layer atlas. Then, we optimize the location of the generated frames via the proposed novel initialization methods and consistency loss.}
    \label{fig:pipeline}
\end{figure*}

\subsection{Vector Sketching Generation}
There are some image-to-vectors methods~\cite{reddy2021im2vec,deepsvg, das2020beziersketch} that can typically produce pixel-aligned results. However, sketching demands a higher level of abstraction and greater continuity between lines. Recent advancements in differential gradient algorithms, such as DiffVG~\cite{diffvg}, have made it feasible to optimize images and even SVG representations within the pretrained CLIP~\cite{clip} space. For example, CLIPDraw~\cite{clipdraw} explores the potential of optimizing SVG images to generate drawings that closely align with text prompts, leveraging the guidance of CLIP. Similarly, Tian and Ha\cite{Tian_Ha_2022} have designed an evolutionary algorithm to abstract images using vector triangles.

Regarding closely related works, CLIPasso~\cite{clipasso} specifically applies the technique of CLIPDraw~\cite{clipdraw} to generate object sketches, and CLIPascene~\cite{clipascene} extends the optimization process to incorporate background elements. Furthermore, similar optimization methods have been employed in the domain of artistic fonts~\cite{iluz2023word} with the diffusion model~\cite{stable-diffusion}. Text-based vector generation has attracted research attention, as evidenced by Wu \emph{et al}.~\cite{wu2023iconshop}, who propose a transformer-based method for generating icons from text auto-aggressively. Different from the vectorization method for images, our approach focuses on generating sketch videos that need to keep temporal consistency, and directly using the image-based method will fall into local minimal due to initialization.

\subsection{Video Editing and Temporal Consistency}
Video editing and stylization have a long history within the computer vision and graphics communities. Various attempts have been made to achieve stylization~\cite{stylit, ebsynth}. However, these methods may face challenges in maintaining tracking consistency. Since frame-wise techniques can generate high-quality stylized images~\cite{gatys2016image, johnson2016perceptual}, it has become a common practice to employ neural networks for reducing temporal inconsistencies as a post-processing step~\cite{dvp, BTSSPP15, lai2018learning, lei2022deep}. Nonetheless, it is important to note that style transfer techniques primarily rely on measuring perceptual distance~\cite{zhang2018perceptual}, which can result in imperfect stylization due to a lack of deep comprehension at the semantic level. Some recent works have shown improved consistency, but often within specific domains, such as portrait videos\cite{fivser2017example, Vtoonify}. For local video editing, layer-atlas-based methods~\cite{layer-atlas, text2live} present a promising approach by allowing video editing on a flattened texture map and generating results through color-wise mapping. \\\\
More recent approaches have explored video editing using diffusion models~\cite{stable-diffusion}. These models offer stronger priors for editing using text. \emph{e.g}., Gen1~\cite{gen1} trains a conditional model for depth and text-guided video generation, allowing on-the-fly appearance editing of generated images. Several methods~\cite{tuneavideo, fatezero, text2video-zero, videop2p} leverage pre-trained text-to-image diffusion models for zero-shot or one-shot video editing. Although current methods have shown promising results for image pixels, there is still a lack of techniques for generating vector video sketches. 
\section{Methods}
\label{sec:method}

Our approach primarily aims to generate the sketch representation of the objects in video, which can be represented through multiple vector strokes. Each stroke is represented as the four points Bézier Curves, where we focus on maintaining both semantic accuracy and temporal consistency. To achieve, this goal, as shown in Figure~\ref{fig:pipeline}, we first decompose a video into a 2D representation using layer atlas~\cite{layer-atlas}~(Sec.~\ref{sec:layer-altas}). Then, we introduce a new framework to optimize the location of these points in Sec.~\ref{sec:optimization}. Finally, we give some potential applications in Sec.~\ref{sec:application}. 

\subsection{Preliminary: Video Decomposition via Layer Atlas}
\label{sec:layer-altas}

\if
Due to the abstract nature of sketches and the absence of ground truth, optimizing each frame's sketches separately based solely on semantic information cannot guarantee the inter-sketch correlation and temporal consistency. In reality, there exists a certain correlation between pixels in a video, meaning that fixed positions have consistent projections across different frames.
\fi

Unlike previous video synthesis and editing tasks~\cite{vid2vid, gen1}, our method focuses on optimizing the positions of discrete points within curves to ensure consistent behavior across frames. Therefore, using image-based frame deflickering methods~\cite{dvp,all-in-one-defliker} directly becomes challenging. To address this, we leverage the trained representation of previous consistent video editing method, \emph{i.e}., Neural Layered Atlas (NLA~\cite{layer-atlas}), to maintain point consistency.

As shown in the left part of Fig.~\ref{fig:pipeline}, the Neural Layered Atlas~(NLA) treats the video as a spatial-temporal 3D volume,where each 3D coordinate within the video is mapped onto the foreground~(or background) 2D $UV$-maps through a Multi-Layer Perceptron~(MLP). 
Additionally, an extra MLP is employed to assign color values to a given 2D UV location. To formalize this, let's consider a video pixel as $p = (x, y, t)$, where $x$ and $y$ are the coordinates of the pixel within the $t$-th frame. NLA generates the 2D UV maps for foreground and background individually as shown in Fig~\ref{fig:pipeline}. It uses $M_f$ and $M_b$ to map 3D coordinate $p$ to a 2D location in foreground UV-map as $(u_f, v_f)$, and in background UV-map as $(u_b, v_b)$ separately. 
The MLP $M_{\alpha}$ indicates the ownership of the points (foreground or background) according to the motion prior and the pre-defined mask losses.
Subsequently, a shared MLP $A$ is trained to map the $(u,v)$ to the corresponding color in RGB space, ensuring that the same pixel in the real world should have the same color.

This comprehensive process facilitates the video reconstruction. After training, the mapping MLPs $M_f$ and $M_b$ indicate the correspondence between points in the video and specific points within the holistic $UV$-map, aligning with our requirements for curve mapping consistency. Our method adheres to the original video decomposition approach of the layer atlas. For more specific information regarding training and loss functions, detailed insights can be found in \cite{layer-atlas}.

\subsection{Differentiable Optimization for Video Sketch}
\label{sec:optimization}

\begin{figure}[t]
    \centering
    \includegraphics[width=\columnwidth]{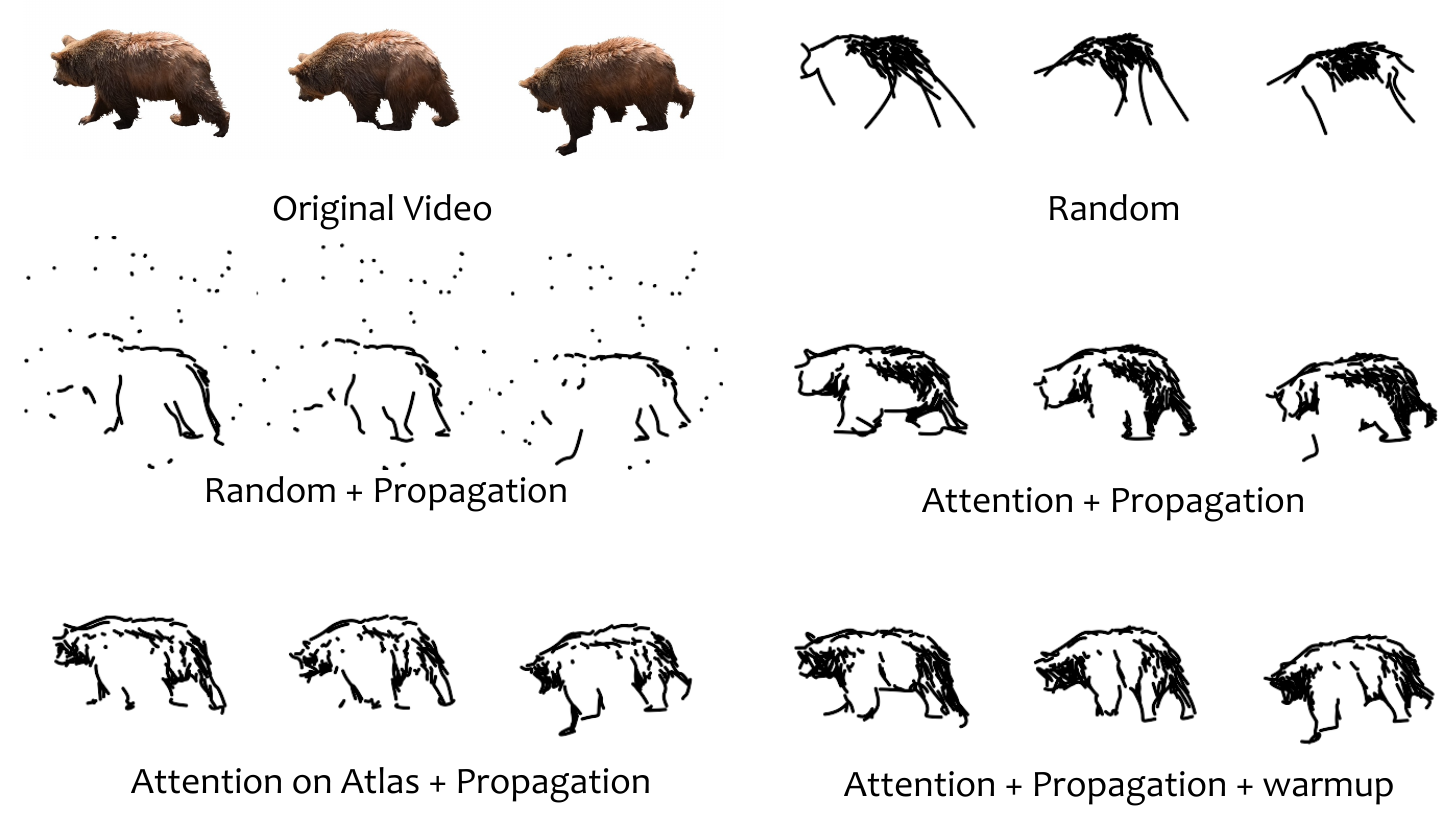}
    \vspace{-2em}
    \caption{The influence of different initialization methods.}
    \label{fig:init}
\end{figure}

In this section, we give the details on how to generate a sketch video in our framework. Formally, given a real video $\{\mathcal{I}_1, ..., \mathcal{I}_T\}$ contains $T$ frames, we define the sketching video $\{\mathcal{S}_1, ..., \mathcal{S}_T\}$ as a set of $N$ strokes $\mathcal{S}_t = \{s_1, ..., s_N\}$ for each frame. 
Each stroke $s_i$ is defined as the two dimensional Bézier curves, where each curve is built via $4$ control points $s_i = \{(x_i, y_i)^k\}_{k=1}^4$. We empirically use the same index $i$ of stroke $s_i$ in different sketches $\mathcal{S}_t$ to represent the same curve across frames and only optimize their positions. All other curve-related attributes are following the previous image sketching method~\cite{clipasso}. We then use a differentiable rasterizer $\mathcal{R}$ from DiffVG~\cite{diffvg} to transform the control points and attributes of Bézier Curves into a Scalable Vector Graphics~(SVG) to represent the final sketch video. So the loss functions $\mathcal{L}$ can be optimized between the real video and the sketches representation. The overall process can be represented as:
\begin{equation}
    \arg\min_{(X,Y)}\sum_{t=1}^T \mathcal{L}(\mathcal{R}(\mathcal{S}_t), \mathcal{I}_{t}),
\end{equation}
where $(X,Y)$ are all the coordinates of the control points $\bigcup_{i=1}^{N}\bigcup_{k=1}^{4}(x_i, y_i)^k$ in the whole sketch video.

Subsequently, we first introduce our method on how to obtain the initial stroke settings on video~(Sec.~\ref{sec:init}). Then, we elaborate on the particulars of rendering points into curves and optimize the entire video~(sec.~\ref{sec:optim}). 

\begin{figure}[t]
    \centering
    \includegraphics[width=\columnwidth]{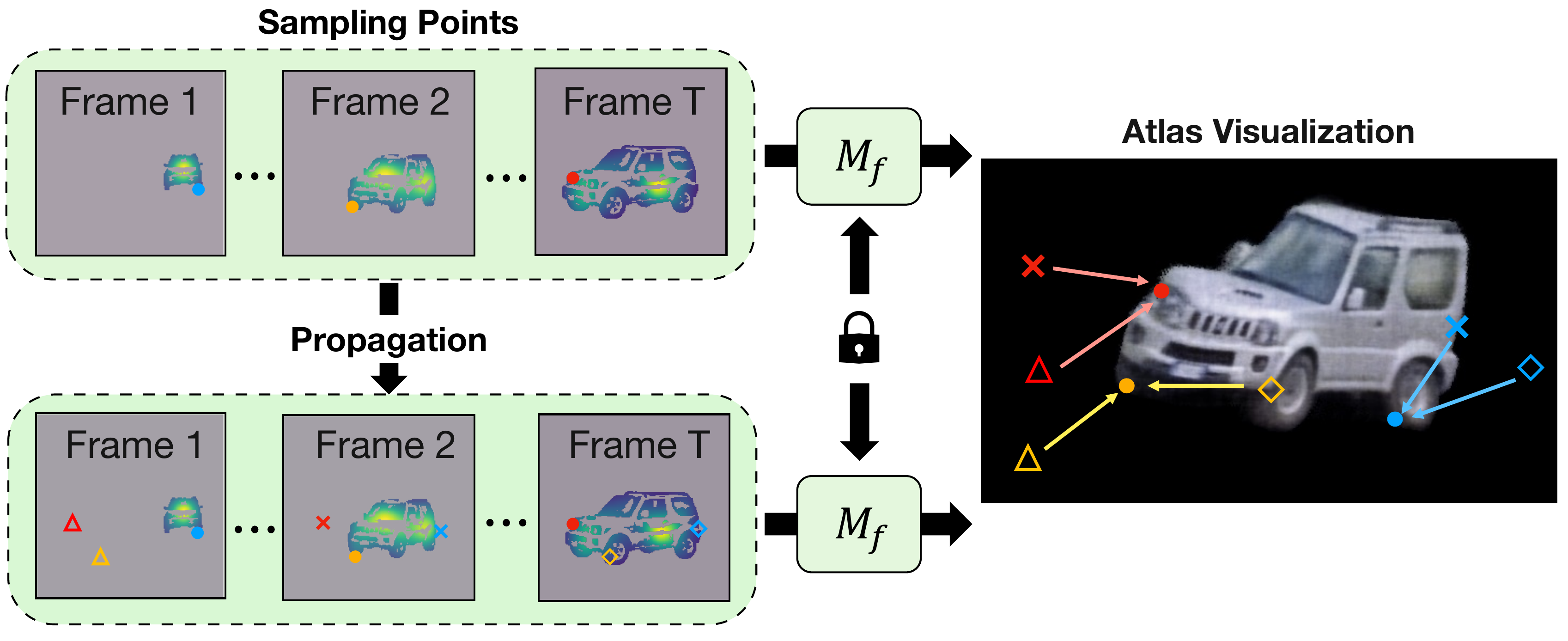}
    \vspace{-2em}
    \caption{\textit{\textbf{Coordinate Initialization.}} The initialized points are warmed up by optimizing the propagated points in different frames $\{$
    {\color{red}$\bigtriangleup$}, {\color{orange}$\bigtriangleup$},  {\color{red}$\times$}, {\color{blue}$\times$}, {\color{blue}$\diamondsuit$},{\color{orange}$\diamondsuit$} $\}$ closer to the sampling points $\{$
    {\color{red}$\bullet$}, {\color{blue}$\bullet$}, {\color{orange}$\bullet$} $\}$ on the atlas~(by mapping MLP $M_{f}$).}
    \vspace{-1em}
    \label{fig:init_method}
\end{figure}

\subsubsection{Strokes Initialization}
\label{sec:init}

The objective function of our abstraction is highly non-convex~\cite{clipasso} since the optimization loss is based on rendered views in pixel differences~\cite{diffvg}. Thus, attempting to optimize the location directly from random initialization often leads to the process getting trapped in local extrema, both in the image and video abstraction~(as in Fig.~\ref{fig:init}). To overcome this challenge, CLIPasso~\cite{clipasso} employs a saliency-guided initialization process, where strokes are sampled from the probability map. This map is generated by multiplying the saliency map via CLIP~\cite{clip} feature with the image's edge map extracted using XDoG~\cite{xdog}, and then subjecting the result to softmax normalization. We argue this kind of initialization is hard to work in processing video since each frame has a variant attention map, where these varying initial points increase the difficulties of the point optimizations. Below, we give our solution step by step.

\noindent\textbf{Point Propagation.} We first aim to generate sparse key points across frames, which are used to represent the entire video. As shown in Figure~\ref{fig:init_method}, to build $T$ frames SVG video where each frame contains $N$ strokes, we first apply salience-guided initialization process~\cite{clipasso} to sample $N$ candidate points in each frame individually. Then, in order to obtain the initial points on video, we resample $N$ points from the set of all $N \times T$ candidate points to build a cross-frame point set $P = \{p_1, ..., p_N\}$. Here, $p_n$ are 3D coordinates $(x_n,y_n,t_n)$, where n indicates the index of control points. Then, these cross-frame points are directly propagated to all frames so that each frame has the same spatial 2D initialization as $\hat{p}^t_n = (x_n, y_n, t)$.

\noindent\textbf{Position Warmup.} Since there can be offsets between the sampled points in different frames~(\emph{e.g.}, the same color markings in various frames in Figure.~\ref{fig:init_method}), we employ an optimization-based approach to alleviate the differences between the propagated locations and the previously sampled positions on the UV maps by adjusting the propagated points only. This alignment is achieved through the use of the pretrained atlas mapping network $M_f$ to optimize the control points across frames on the atlas:
\begin{equation}
    \mathcal{L}_{warmup} = \sum_{t=1}^T\sum_{n=1}^N|| M_f(\hat{p}_n^t)) - M_f(p_n)) ||_1,
\end{equation}
In this situation, $t$ represents the frame index, $n$ signifies the control point index, $p$ denotes the sampled points, and $\hat{p}$ signifies the propagated points.

In practice, we warm up 300 iterations. This strategy helps us to find the most suitable initialization points. As shown in Fig.~\ref{fig:init}, random initialization (or solely attention-based methods~\cite{clipasso} with similar results) yields poor performance due to the lack of correspondence between the same index points across the video. While propagating the points across different frames helps establish correspondence, it can still lead to a loss of focus on the object. Different attention strategies yield varying results, with frame-wise attention outperforming the attention map based on the atlas. This is because the atlas tend to exhibit distortions compared to the more realistic frames. Ultimately, the performance is further enhanced and becomes more accurate with the inclusion of a warm-up phase in the entire initialization process.

\begin{figure}[h]
    \centering
    \includegraphics[width=\columnwidth]{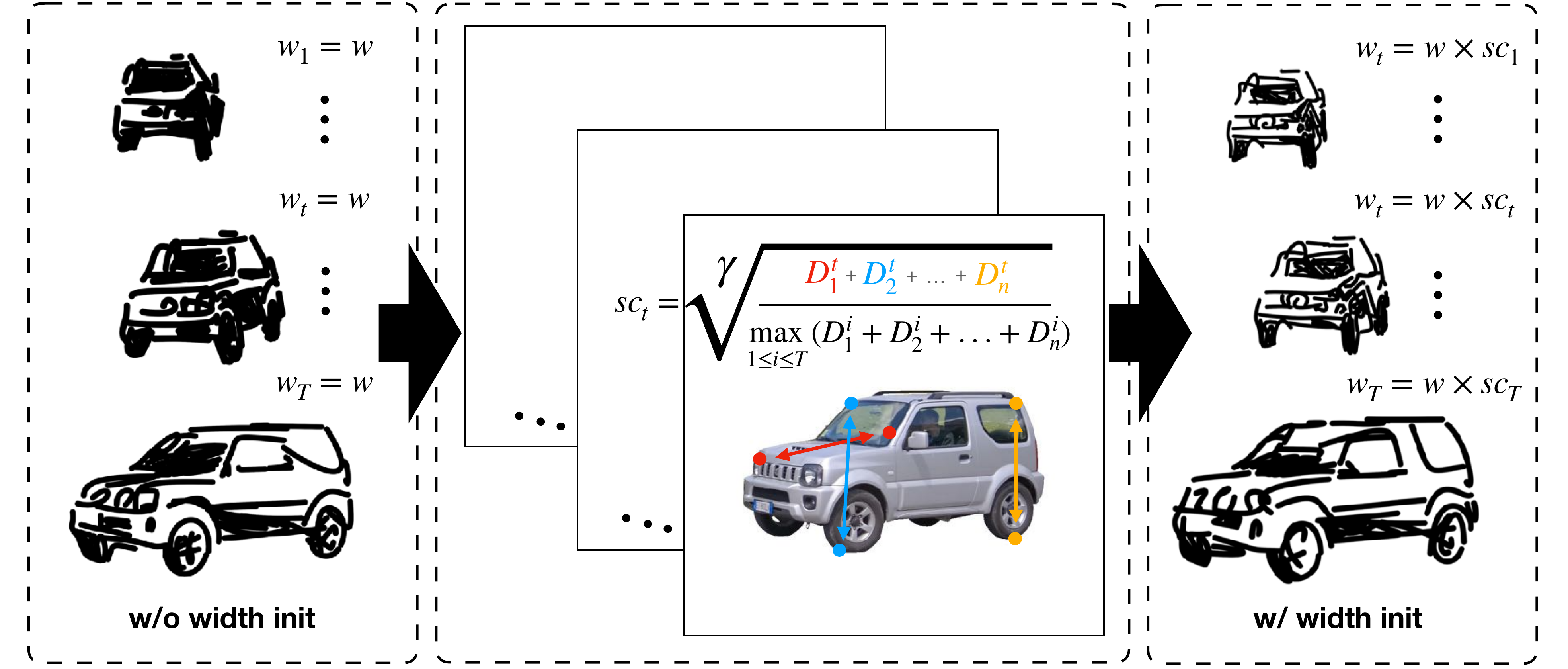}
    \vspace{-2em}
    \caption{The width $w_t$ of strokes in $t$-th frame is scaled by the variable $sc_t$, with initialized width $w$. The variable $
    \gamma$ controls the contrast ratio of scale (default value is 3).}
    \label{fig:width_init}
\end{figure}

\noindent\textbf{Curve Width Initialization.} 
Additionally, we also initiate the curve widths, considering that generated videos can involve significant motions and varying scales. Since more accurate stroke thickness enhances the representation of contours, instead of solely relying on the mask's scale, we leverage the object's distortion to estimate the appropriate scale. In detail, as depicted in Figure.~\ref{fig:width_init}, we randomly pair points in each frame to calculate the differences between points~(here, we use all warmed-up points to create pairs). By computing the $\gamma$ root of the weighted pair-wise differences on frame $t$~(with $\gamma$ defaulted to 3 as a measure of distortion in the 3D world), we derive a scale factor $sc_t$ and curve width $w_t$:
\begin{equation}
    sc_t = \sqrt[\gamma]{\frac{\sum_{n=1}^N D_n^t}{\mathop{\max}\limits_{1 \leq i \leq T}(\sum_{n=1}^N D^{i}_n)}},
\end{equation}
\begin{equation}
    w_t = sc_t \times w,
\end{equation}
where $w$ is the default width. Consequently, more distant objects are depicted with finer lines to capture intricate details and scale, while closer objects are outlined with thicker lines. This strategy enhances the quality of abstraction, particularly in scenarios involving substantial object motion.

\subsubsection{Curves Optimization}
\label{sec:optim}

Following a well-executed stroke initialization for the video, our aim is to optimize the curves based on the positions of their points. This optimization needs to ensure that the resulting sketch video maintains not only semantic similarities with the input video but also consistency across frames.

Firstly, we follow image-based sketching methods~\cite{clipasso} to use the semantic-aware loss~$\mathcal{L}_{semantic}$ to measure the differences between the generated points and the original video. $\mathcal{L}_{semantic}$ is based on the differences in multi-scale perception features extracted from the pre-trained CLIP image encoder~\cite{clip}. Since CLIP is trained on a larger-scale dataset to align the text and image information through contrastive learning, the semantic information can be well-aligned. Formally, for the control points $\mathcal{S} = \{s_1,...,s_N\}$ where $s_i = \{(x_i, y_i)^k\}_{k=1}^4$ on each SVG, we optimize the positions of the control points based on the difference between the generated SVG and the actual input image $\mathcal{I}$ via the $l$-th layer of pretrained CLIP $\Phi$:
\begin{equation}
    \mathcal{L}_{semantic} = \sum_{l\in [3,5,9]}\sum_{t=1}^T|| \Phi_l(\mathcal{R}(\mathcal{S}_t)) - \Phi_l(\mathcal{I}_t)) ||_1,
\end{equation}
Here, $T$ represents the total number of frames, and $\mathcal{R}$ is the differentiable rasterizer as introduced above, 

\begin{figure}[ht]
    \centering
    \includegraphics[width=\columnwidth]{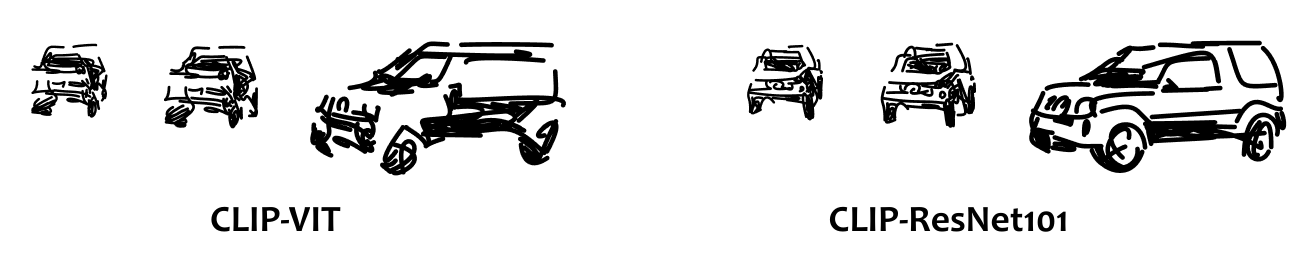}
    \vspace{-2em}
    \caption{The differences in the choice of different semantic losses.}
    \label{fig:clip}
\end{figure}

We also investigate the impact of variants of the CLIP encoders, specifically, the ViT~\cite{vit} based and ResNet101-based models~\cite{resnet}. The ResNet101-based CLIP model exhibits better performance with more local structures, while the ViT-based methods focus more on global features. Consequently, we default to using the ResNet101-based CLIP model, as depicted in Figure~\ref{fig:clip}.

Besides, for our video sketching, we design a novel consistency loss to measure the consistency between the generated SVG videos. The most naive approach is to maintain control over the corresponding offsets to guarantee temporal consistency via optical flow~\cite{raft,vid2vid}. While optical flows are typically dense, they might suffer from potential errors, \emph{e.g}., forward-backward consistency, and cumulative errors. Therefore, we employ the trained atlas network to obtain a panoramic view of the video. This allows each point in every frame of the video to be associated with a consistent global position on the atlas. By ensuring that related points and lines have consistent global positions, we can maintain the continuity of the video. Specifically, we use the pre-trained mapping network $M_f$ from the layer atlas as shown in Figure~\ref{fig:pipeline}, where for each control point $\hat{p}_n^{t_1}$ and its neighboring $\hat{p}_n^{t_2}$ of the same index $n$ cross frames, where the $\mathcal{L}_{consistency}$ can be written as:
\begin{equation}
    \mathcal{L}_{consistency} = \sum_{\hat{p}'\in \mathcal{N}(\hat{p})}\sum_{\hat{p} \in P}|| M_f(\hat{p})) - M_f(\hat{p}') ||_1,
\end{equation}
After optimization, the same index $n$ point $\hat{p}_n$ cross frames will have a similar location in the 2D $UV$-map and it keeps consistency over frames.

Overall, the loss function can be written as:
\begin{equation}
    \mathcal{L} = \omega_1 \mathcal{L}_{semantic} + \omega_2 \mathcal{L}_{consistency},
\end{equation}
where $\omega_1$ and $\omega_2$ are used to control the extent of semantics and consistency between sketches, respectively. We experimentally set $\omega_1 = 200.0 $ and $\omega_2 = 3.0 $.   

\subsection{Applications}
\label{sec:application}

\noindent\textbf{SVG Editing.} Because the generated video is in the form of SVG, our method has the ability to change the colors of all lines or fill specific lines within the video to create richer details in the visual content. As shown in Fig.~\ref{fig:teaser}, we can resize the canvas, paint all objects in orange, and fill the enclosed lines.

\noindent\textbf{Video Editing and Doodling.} Our method can also be employed for video editing and doodle creation. As shown in Fig.~\ref{fig:teaser}, after generating the SVGs, we can use inpainting techniques\cite{lama} to restore the original video and integrate the newly generated content into it. Additional video results are presented in the supp. video.
\section{Experiments}

\label{sec:exp}




Our method is evaluated on DAVIS dataset~\cite{davis}, which provides foreground annotations for each frame. We also evaluate the proposed method on some self-collected datasets, utilizing the video matting method~\cite{rvm} to generate the foreground videos. In each case, we only used the first 50 frames in the video to generate the results for a fair comparison. Subsequently, we decompose the video into an atlas. For the optimization of sketching videos, we utilize the Adam optimizer with a learning rate of 1.0, following the approach of CLIPasso~\cite{clipasso}. On average, optimizing a video sketch takes approximately 29 minutes, consuming around 19.5GB on a single NVIDIA GeForce RTX 3090 GPU.

\begin{figure*}[t]
     \centering
     \begin{subfigure}[b]{\textwidth}
         \centering
        \includegraphics[width=\textwidth]{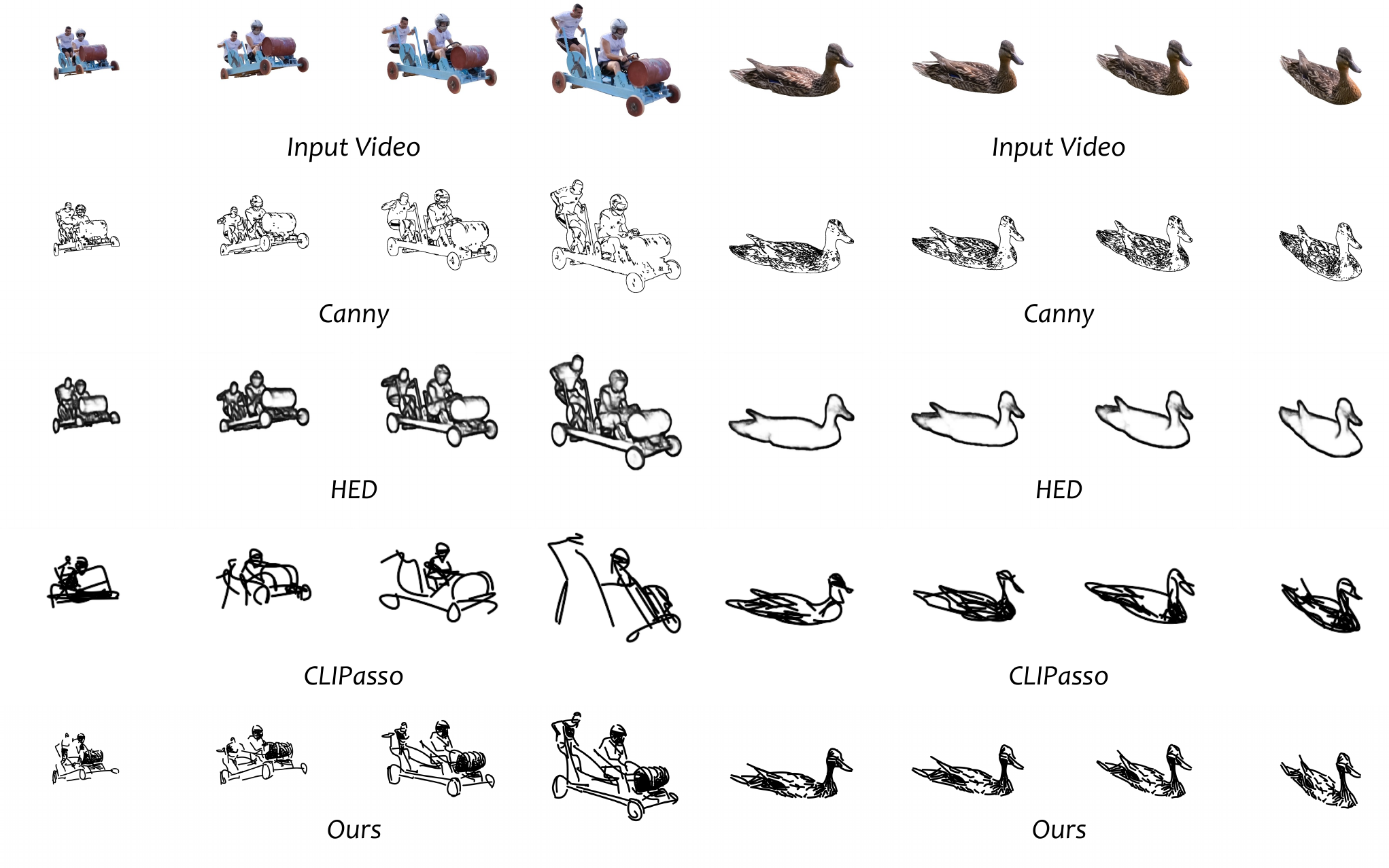}
     \end{subfigure}
     \vspace{-2em}
    \caption{Comparisons with our methods and the states-of-the-art methods frame-wise methods, \ie, frame-wise CLIPasso~\cite{clipasso} and edge detection methods~(canny~\cite{canny} and HED~\cite{hed}) on different frames of the videos.}
    \vspace{-1em}
    \label{fig:comparsion}
\end{figure*}

\subsection{Compare with state-of-the-art methods}

As no previous video sketching methods exist, we employ some image sketching and edge detection techniques for comparison. For image abstraction, in particular, we compare with frame-by-frame CLIPasso~\cite{clipasso}. Regarding edge detection, we assess against traditional edge detection techniques, namely Canny~\cite{canny}, as well as the deep learning-based method, HED~\cite{hed}. As demonstrated in Figure~\ref{fig:comparsion}, in comparison to CLIPasso, our approach exhibits improved temporal consistency while effectively preserving semantic information. Concerning the edge detection methods, our proposed technique showcases superior semantic-aware abstraction. Further insights and video comparisons can be found in the supplementary video.

\begin{table}[b]
\centering
\begin{tabular}{l|cccc}
\toprule  
& Temporal $\uparrow$  & Abstraction $\uparrow$  & Overall $\uparrow$ \\
\hline
Canny  &2.65  &2.58 &2.51   \\
HED  &2.67  &2.30 & 2.60  \\
\hline
CLIPasso &2.13  &2.41 &2.31  \\
Ours & 2.58  & \textbf{2.74} & \textbf{2.71}    \\
\bottomrule
\end{tabular}
\vspace{-1em}
\caption{The mean opinion of the user studies.}
\label{tab:user_study}
\end{table}

Recognizing the absence of appropriate metrics for numerically evaluating sketching, we conduct user studies to assess the performance of the generated video. In detail, for each of the 12 clips, we provide generated videos created by different methods for comparison. We then invite 25 individuals to rank the resulting videos in terms of semantic alignment (abstraction), temporal consistency (temporal), and overall quality (Overall), respectively. The order of the generated videos is shuffled, and participants score the results from 1 to 4 based on their ranking. As presented in Table ~\ref{tab:user_study}, our method receives more favorable feedback from users when compared to the baseline CLIPasso~\cite{clipasso}, demonstrating improvements across temporal consistency, semantic abstraction, and overall quality. Moreover, it's noteworthy that our approach achieves enhanced semantic results with scores on par with those of edge detection methods in terms of temporal consistency.

\begin{figure}[t]
    \centering
    \includegraphics[width=\columnwidth]{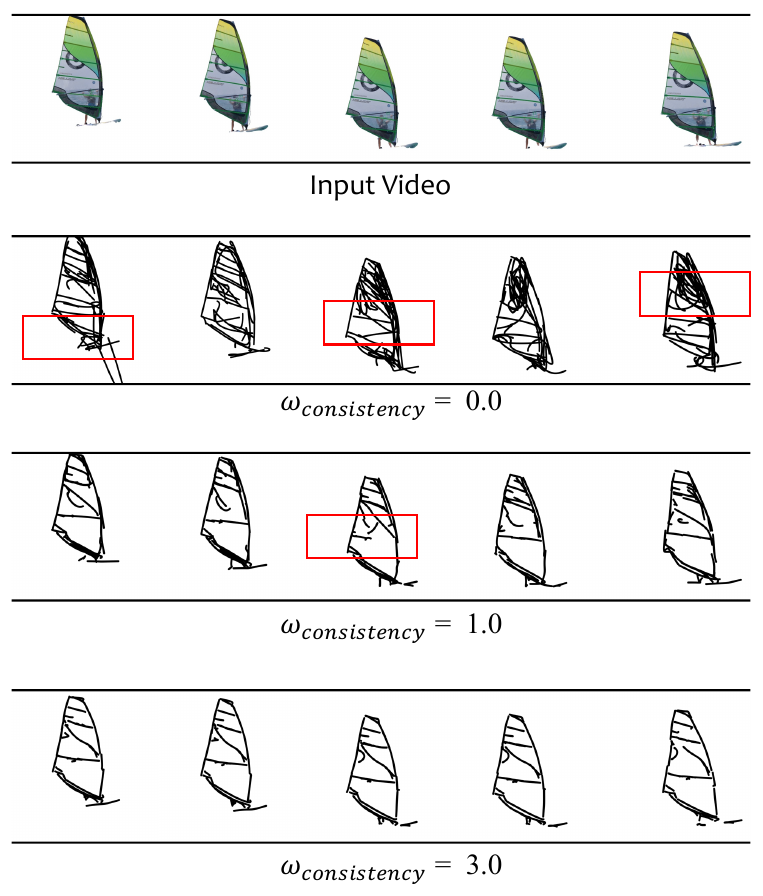}
    \vspace{-2em}
    \caption{The importance of the proposed consistency loss.}
    \label{fig:consist_param}
    \vspace{-1em}
\end{figure}

\subsection{Ablation Studies}


\begin{figure}[t]
    \centering
    \includegraphics[width=\columnwidth]{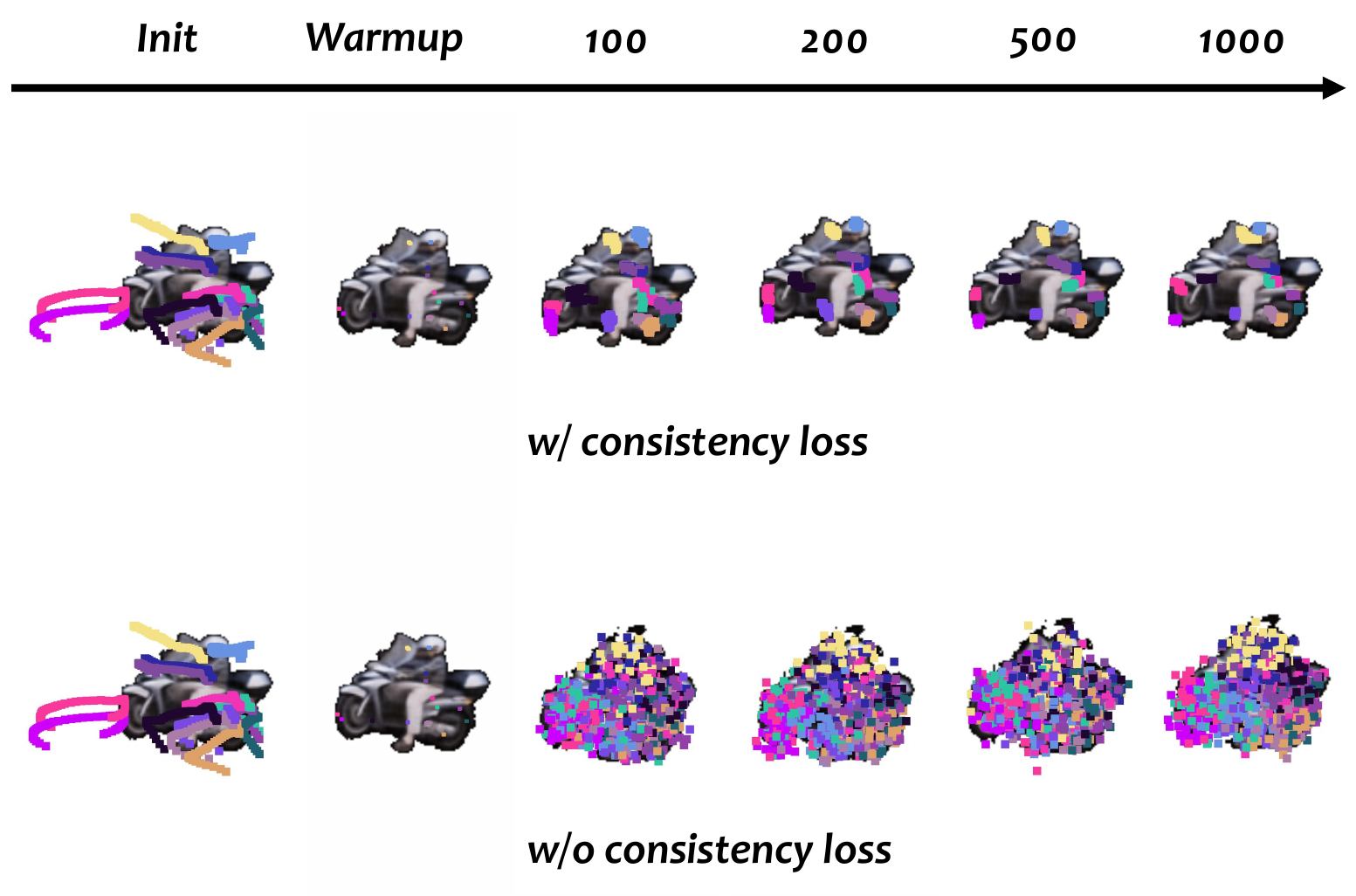}
    \vspace{-2em}
    \caption{\textbf{\textit{Point Visualization}}. To further support the effectiveness of the proposed consistency loss on the atlas, we visualize the location of the same curves~(same color) on different frames during optimization. Kindly take note that the warmup points have been gathered, and it is recommended to view them with zoom in.}
    \vspace{-1em}
    \label{fig:atlas_show}
\end{figure}

\noindent\textbf{Optimization Visualization.} We introduce a novel consistency loss based on the trained atlas network. Here, we visualize the points on the global UV map (atlas) to gain a clearer understanding. As illustrated in Fig~\ref{fig:atlas_show}, points with the same index across multiple frames are represented in the same color. During the optimization process, the control points from different frames are appropriately positioned at the same points after the warming-up phase. Subsequently, we optimize these points using both the consistency loss and the semantic loss to ensure performance in terms of semantic alignment and coherence.

\noindent\textbf{Consistency Weights.}
The novel consistency loss maintains temporal domain consistency through the trained atlas representation. In our ablation study, we investigate this loss using different values. As depicted in Fig.~\ref{fig:consist_param}, when the consistency loss is excluded (\ie, $\omega_{consistency} = 0 $), the generated sketch exhibits distinct representations across frames. As we increment this parameter, the optimized results demonstrate enhanced stability.

\begin{figure}[t]
    \centering
    \includegraphics[width=\columnwidth]{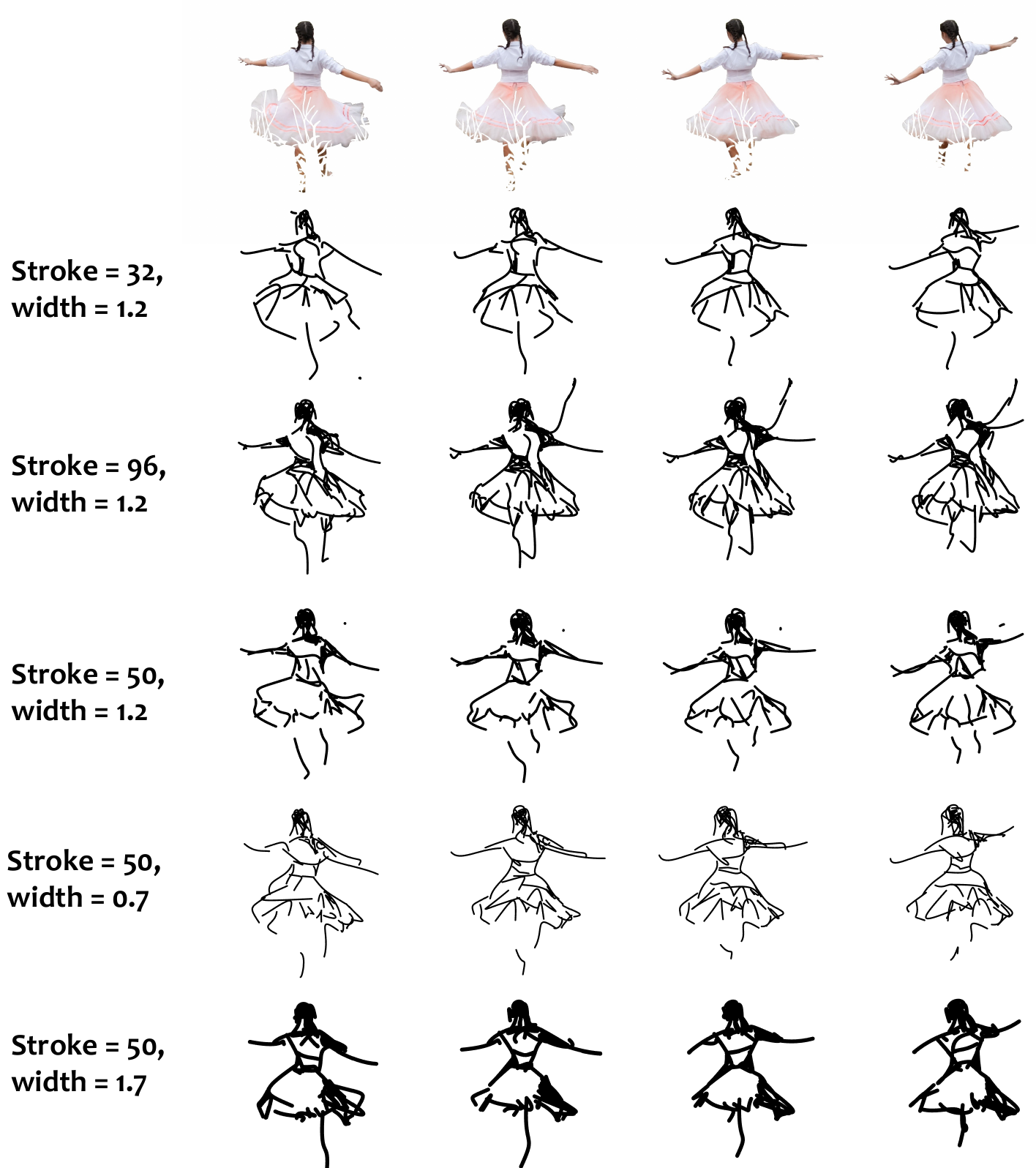}
    \vspace{-2em}
    \caption{The ablations on the width and stroke numbers.}
    \label{fig:number_of_sketching}
\end{figure}

\noindent\textbf{Strokes Number and Width.} We also show the SVG using different numbers and widths of the default strokes. Increasing the number of paths, as illustrated in Fig.~\ref{fig:number_of_sketching}, leads to the generation of sketches that capture more intricate details. For instance, the details of the dress and its movements become more prominent and well-defined, which reduces the level of abstraction. Likewise, adjusting the stroke width yields similar effects. Decreasing the stroke width during optimization emphasizes finer details within the strokes rather than the overall structure, resulting in a less abstract representation.

\subsection{Limitation}
\begin{figure}[t]
    \centering
    \includegraphics[width=\columnwidth]{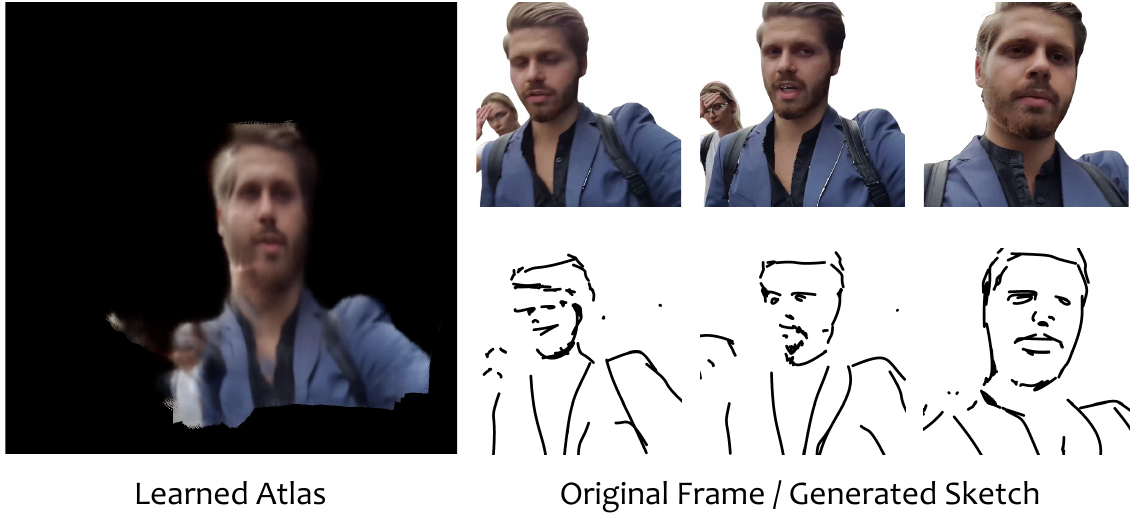}
    \vspace{-2em}
    \caption{\textit{\textbf{Limitations}}. When the motion is large and the object is complex,  the generated atlas may not be accurate, causing temporal inconsistency.}
    \vspace{-1em}
    \label{fig:limitation}
\end{figure}

While our method excels at generating coherent and semantically rich videos, the limitations of the trained layer atlas~\cite{layer-atlas} constrain the quality of the sketches. For instance, it faces challenges when representing the motion of non-rigid bodies. Additionally, during cases of self-occlusion, the generated abstract sketches may contain errors, often appearing as improper turns (as observed in the supplementary video). Furthermore, the generated sketches may exhibit some texture artifacts when the video undergoes significant motion or involves complex foreground elements. As depicted in Fig.~\ref{fig:limitation}, our proposed method exhibits artifacts when the male face shakes violently, and the atlas struggles to accurately represent the woman in the image. While some of these issues can be mitigated by subdividing the video into smaller sections and employing more layers, the accuracy of segmentation and the computational cost associated with multi-layer approaches remain challenges.

\subsection{Conclusion}

We present an optimization-based approach for generating sketch videos that maintain both semantic and temporal consistency. Our method facilitates the creation of sketching videos with the appropriate level of abstraction. Essentially, it includes a novel initialization technique for acquiring well-initialized points and a distinctive consistency loss derived from self-supervised video decomposition. These innovations empower us to craft sketch videos using simple Bézier curves. Since the resulting videos are composed using Scalable Vector Graphics, our proposed methods offer versatile applications in video editing and doodling, accommodating various sizes while preserving intricate details.


\end{document}


\title{\paperTitle \\ \textit{Supplemental Material}}
\author{\authorBlock}
\maketitle

\appendix
\label{sec:appendix}





\section{Implementation Details}
\label{sec:Implementation Details}
\noindent \textbf{Pseudo algorithm code} Our full algorithm is shown in Algorithm~\ref{alg:real_image_editing} and Algorithm~\ref{alg:attention_fusion}. Algorithm~\ref{alg:real_image_editing} presents the overall framework of our inversion and editing, as visualized in the left of Fig. \textcolor{red}{1} in the main paper.
Algorithm~\ref{alg:attention_fusion} shows that the cross-attention is fused based on a mask of the edited words, and the self-attention is blended using a binary mask from thresholding the cross-attention (the right of Fig.~\textcolor{red}{1} in the main paper).

\noindent \textbf{Hyperparameters Tuning}. There are mainly three hyperparameters in our proposed designs:
 \\
 - ${t}_s \in [1, T]$: Last timestep of the self-attention blending. Smaller ${t}_s$ fuses more self-attention from inversion to preserve structure and motion.
 \\
 - ${t}_c \in [1, T]$: Last timestep of the cross attention fusion. Smaller ${t}_c$ fuses more cross attention from inversion to preserve the spatial semantic layout.
 \\
 - $\tau\in [0, 1]$: Threshold for the blending mask used in shape editing. Smaller $\tau$ uses more self-attention map from editing to improve shape editing results.

 In \textbf{style} and \textbf{attribute} editing, we set ${t}_s=0.2T$, ${t}_c=0.3T$, $\tau=1.0$ to preserve most structure and motion in the source video.
 In \textbf{shape} editing, we set ${t}_s=0.5T$, ${t}_c=0.5T$, $\tau=0.3$ to give more freedom in new motion and 3D shape generation.

\section{Demo Video}
\label{sec:demo_video}
\noindent we provide a detailed demo video to show:

\noindent\textbf{Video Results} on style, local attribute, and shape editing to validate the effectiveness of the proposed method.

\noindent\textbf{Method Animation} to provide a better understanding of the proposed method.

\noindent\textbf{Baseline Comparisons} with previous methods in video.

\noindent\textbf{More Promising Applications} We have shown the effectiveness of the proposed method in the main paper for style, attribution, and shape editing. In the demo video, we also show some potential applications of the proposed method, including (1) object removal by removing the word of the target object in the source prompt and mask the self-attention of the corresponding area using its cross attention, (2) video enhancement by adding the specific prompt~(\eg, `high-quality', `8K') in the target editing prompt.
~\label{sec:Additional application}

\newpage
\section{Limitation and Future Work}
Our zero-shot editing is not good at new concept composition or generation of very different shapes. For example, the result of editing `black swan' to `yellow pterosaur' in Fig~\ref{fig:limitation} is unsatisfactory. This problem may be alleviated using a stronger video diffusion model, which we leave to future work.~\cite{qi2023fatezero}
\label{sec:limitation}
\begin{figure}[t]
\centering

\newcommand{\imwidth}{0.45\textwidth}
\begin{tabular}{@{}c@{}}
\parbox{\imwidth}{\includegraphics[width=\imwidth, ]{figs/imgs/limitation-cropped.pdf}}
\\
{
{black swan} $\xrightarrow{}$ \textcolor{red}{ yellow pterosaur}.
}
\end{tabular}
\caption{limitation of our zero-shot editing.
}
  \label{fig:limitation}
\end{figure}%

{\small
\bibliographystyle{ieee_fullname}
\bibliography{11_references}
}